\documentclass[letterpaper]{article} 
\usepackage{aaai24}  
\usepackage{times}  
\usepackage{helvet}  
\usepackage{courier}  
\usepackage[hyphens]{url}  
\usepackage{graphicx} 
\usepackage{natbib}  
\usepackage{caption} 
\usepackage{algorithm}
\usepackage{algorithmic}
\usepackage {booktabs}
\usepackage{multirow}
\usepackage{bm}
\usepackage{subcaption}
\usepackage{pifont}
\usepackage{amsmath}
\usepackage{amssymb}
\usepackage{color}
\usepackage{colortbl}
\usepackage{soul}
\usepackage{newfloat}
\usepackage{listings}
\DeclareCaptionStyle{ruled}{labelfont=normalfont,labelsep=colon,strut=off} 
\floatstyle{ruled}
\newfloat{listing}{tb}{lst}{}
\floatname{listing}{Listing}
\title{Test-Time Domain Adaptation by Learning Domain-Aware Batch Normalization}
\author{
    Yanan Wu$^{1,2}$\thanks{The authors contributed equally to this work.}, 
    Zhixiang Chi$^{3}$\footnotemark[1], 
    Yang Wang$^{4}$, 
    Konstantinos N. Plataniotis$^3$,
    Songhe Feng$^{1,2}$\thanks{Corresponding Author}
}
\affiliations{

    $^{1}$Key Laboratory of Big Data \& Artificial Intelligence in Transportation, \\ Ministry of Education, Beijing Jiaotong University, Beijing, 100044, China\\
$^{2}$School of Computer and Information Technology, Beijing Jiaotong University, Beijing, 100044, China\\
$^{3}$The Edward S Rogers Sr. ECE Department, University of Toronto, Toronto, M5S3G8, Canada\\
$^{4}$Department of Computer Science and Software Engineering, Concordia University, Montreal, H3G2J1, Canada\\
{\tt\small \{ynwu0510,shfeng\}@bjtu.edu.cn}, {\tt\small zhixiang.chi@mail.utoronto.ca} \\{\tt\small yang.wang@concordia.ca}, {\tt\small kostas@ece.utoronto.ca}
}

\usepackage{bibentry}

\begin{document}

\maketitle

\begin{abstract}
Test-time domain adaptation aims to adapt the model trained on source domains to unseen target domains using a few unlabeled images. Emerging research has shown that the label and domain information is separately embedded in the weight matrix and batch normalization (BN) layer. Previous works normally update the whole network naively without explicitly decoupling the knowledge between label and domain. As a result, it leads to knowledge interference and defective distribution adaptation. In this work, we propose to reduce such learning interference and elevate the domain knowledge learning by only manipulating the BN layer. However, the normalization step in BN is intrinsically unstable when the statistics are re-estimated from a few samples. We find that ambiguities can be greatly reduced when only updating the two affine parameters in BN while keeping the source domain statistics. To further enhance the domain knowledge extraction from unlabeled data, we construct an auxiliary branch with label-independent self-supervised learning (SSL) to provide supervision. Moreover, we propose a bi-level optimization based on meta-learning to enforce the alignment of two learning objectives of auxiliary and main branches. The goal is to use the auxiliary branch to adapt the domain and benefit main task for subsequent inference. Our method keeps the same computational cost at inference as the auxiliary branch can be thoroughly discarded after adaptation. Extensive experiments show that our method outperforms the prior works on five WILDS real-world domain shift datasets. Our method can also be integrated with methods with label-dependent optimization to further push the performance boundary. Our code is available at \url{https://github.com/ynanwu/MABN}.

\end{abstract}

\section{Introduction}
Deep models achieve astonishing performance due to the matched training and testing data distributions~\cite{choi2018stargan,wu2021gm}. However, such assumption is vulnerable in the real world as it is impossible to collect training data to cover the universal distribution. Therefore, unseen distributions at inference lead to degenerate performance stemming from distribution shift~\cite{geirhos2018generalisation}.  

\begin{figure}[t]
\centering
\centerline{\includegraphics[scale=0.99]{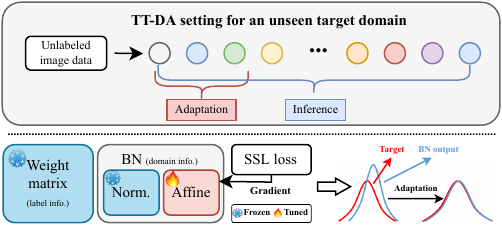}}
\caption{\textbf{Top:} Illustration of TT-DA setting. Given an unseen target domain at test-time, a few unlabeled data are used for adaptation and the adapted model is then used for inference. \textbf{Bottom:} To maximize the extraction of domain knowledge from unlabeled data, we only update the two affine parameters in BN using the label-independent self-supervised loss. The goal is to enforce the adapted affine parameters to correct the feature distribution towards target domain.} 
\label{fig1}
\end{figure}

Unsupervised domain adaptation (UDA) is a line of research to mitigate the distribution shift by incorporating mutual dependence of labeled source data and unlabeled target data~\cite{wang2018deep,ganin2016domain,shu2018dirt,long2018conditional}. Clearly, it is impractical to repetitively perform UDA for every unseen target domain. In contrast, domain generalization (DG) aims to overcome this limitation by aspiring to train a model on source data that can effectively generalize across unseen target domains~\cite{muandet2013domain,seo2020learning}. Nevertheless, it is unrealistic to expect a universal model to handle all diverse unseen domains. To address such limitation, pioneer works such as ARM~\cite{zhang2021adaptive} and Meta-DMoE~\cite{zhong2022meta} propose to introduce an additional phase where the model is adapted to \textit{each} target domain using a few unlabeled data prior to inference. We refer to this scenario as test-time domain adaptation (TT-DA) as shown in the top of Fig.~\ref{fig1}. The motivation behind TT-DA is that a small number of unlabeled data in a target domain are easy to obtain (e.g., images obtained during camera calibration) and they provide a cue of the underlying distribution of the target domain~\cite{zhang2021adaptive}. 

Adapting a model with a limited amount of unlabeled data remains a challenge. ARM tackles the problem by meta-learning adaptive modules for unsupervised adaptation~\cite{finn2017model}. Its main limitation is that the inner and outer loop optimizations are performed on the \textit{same} batch of unlabeled data. Intuitively, the adaptive modules are optimized to only adapt to a \textit{single} batch instead of the broader data distribution. Unfortunately, this can impede effective generalization. Under more challenging real-world benchmarks, ARM even sometimes falls behind the baseline of Empirical Risk Minimization (ERM) as shown in the WILDS leaderboard~\cite{koh2021wilds}.
In contrast, Meta-DMoE enforces the adapted model to generalize on a disjoint query set during training to improve the overall performance. However, Meta-DMoE formulates the adaptation as a knowledge distillation process by querying the target-related knowledge from a group of teacher models trained on source domains. The adaptation is inherently bounded by the teacher models. In addition, the size of the teacher ensemble grows with the number of source domains,  amplifying computational demands and considerably slowing down the adaptation process. It is worth noting that both ARM and Meta-DMoE lack a deliberate decoupling of knowledge between domain and label. This potential overlap of knowledge can introduce interference, rendering the models susceptible to performance degradation~\cite{standley2020tasks}.

In this work, we propose a simple yet effective solution to enhance the refinement of domain knowledge acquisition, achieved by decoupling the label-related knowledge. Our work is partly inspired by the observation that the weight matrix tends to encapsulate label information, while domain-specific knowledge is embedded within the BN layer~\cite{li2016revisiting}. We propose a strategic manipulation of the BN layer to optimize the acquisition and transference of domain-specific knowledge. The BN layer normalizes the input feature followed by re-scaling and shifting using two affine parameters. However, the normalization statistics computed for the target domain under TT-DA can be unstable since we only have a small batch of examples from the target domain. Instead, we propose to only adapt the two affine parameters while directly using the normalization statistics learned from source domains during training. 
The intuition is that the features will first be normalized towards the source distribution, but the adapted affine parameters aim to pull the normalized feature towards the target distribution, as shown in the bottom of Fig.~\ref{fig1}. 
Without the ambiguity caused by unstable statistic estimation, optimization of learning domain knowledge at the adaptation step is much more stable. On the other hand, such a strategy also keeps the same computational cost during inference without incurring additional operations~\cite{du2020metanorm} or parameters~\cite{bronskill2020tasknorm} that are laborious to learn better statistics. Furthermore, to generate a domain-oriented supervision signal on the unlabeled data, we construct an auxiliary branch and employ class-independent self-supervised learning (SSL)~\cite{grill2020bootstrap,liang2022self}. 

Overall, we perform two-phase training. In the first phase, we train the whole model to learn the label knowledge and normalization statistics by mixing all the source data. To endow the affine parameters with the adaptive capabilities to new domains, in the second phase, we employ a bi-level optimization as in meta-learning. Concretely, we treat every source domain as one ``task'' and use few-shot unlabeled images to update the affine parameters (while keeping other model parameters fixed) via an auxiliary branch. To ensure that the optimization of the affine parameters is aligned with the main task, we define the meta-objective by evaluating the adapted affine parameters on a disjoint set in the task. Note that the affine parameters are optimized at the meta-level to act as an initialization for adaptation~\cite{finn2017model}. Through such a learning paradigm, our model is learned 
 to effectively adapt to a domain using unlabeled data and use the adapted model to perform inference. We dub our method as \textbf{M}eta-\textbf{A}daptive \textbf{BN} (MABN). Our contributions are summarized as follows:
\begin{itemize}
    \item We propose a simple yet effective unsupervised adaptive method that is tailored for TT-DA. We adapt only the affine parameters via a self-supervised loss to each target domain to elevate the domain knowledge learning.
    \item We employ a bi-level optimization to align the learning objective with the evaluation protocol to yield affine parameters that are capable of adapting domain knowledge.
    \item We conduct extensive experiments to show that our method is more effective in learning the domain knowledge. Thus, our domain-adapted model can be seamlessly integrated with the entropy-based TTA method (e.g. TNET~\cite{wang2020tent}) where the optimization is more toward label knowledge. 
    \item We surpass ARM and Meta-DMoE by 9.7\% and 4.3\% of Macro F1 on WILDS-iWildCam. We achieve superior performance on five real-world domain shift benchmarks in WILDS with both classification and regression tasks. 
\end{itemize}

\section{Related Work}

\paragraph{Domain shift.} 
\textit{Unsupervised Domain Adaptation} (UDA) addresses the distribution shift by jointly training on the labeled source and unlabeled target data. Popular methods include aligning the statistical discrepancy across various distributions~\cite{peng2019moment} and developing common feature space via adversarial training~\cite{pei2018multi, long2018conditional}. Recently, \textit{source-free} UDA methods have been proposed to allow the absence of source data, such as generative-based methods~\cite{li2020model, qiu2021source}. However, both settings are unrealistic. They have to access sufficiently large unlabeled target datasets, and their objective is to achieve high performance on that particular target domain. \textit{Domain Generalization} (DG) is another line of research that learns a generic model from one or several source domains, and expects it to generalize well on unseen target domains \cite{zhou2020learning, lv2022causality}. Both types of domains cannot be accessed simultaneously.  
However, deploying such model to all unseen target domains fails to explore domain specialty and often yields inferior solutions. 

\begin{figure*}[t]
\centering
\includegraphics[width=1.8\columnwidth]{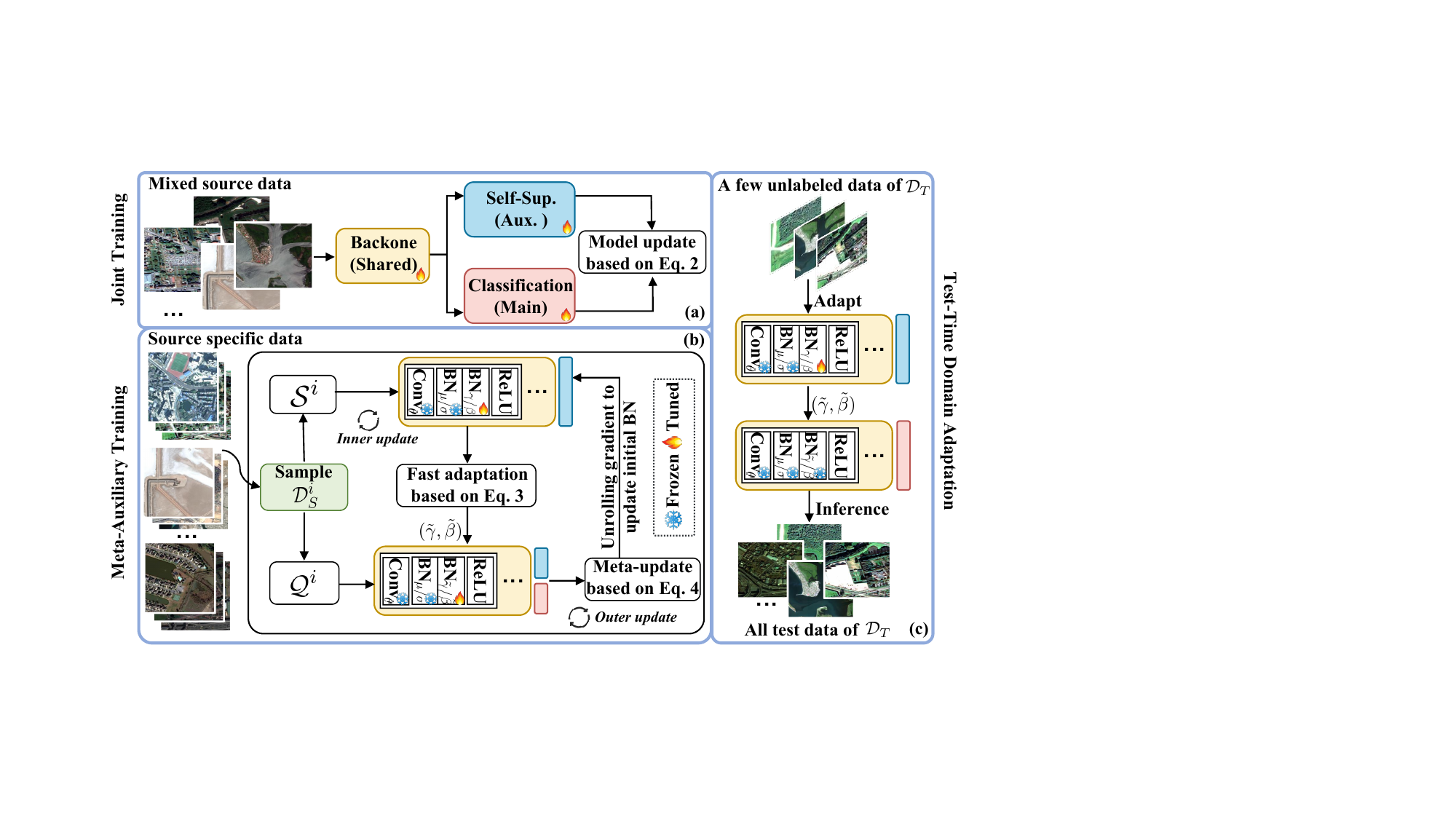} 
\caption{{\bf Overview of the proposed MABN}. In the joint training stage (a), we train the entire network to learn both label knowledge and normalization statistics by mixing all the source data and performing joint training. During the meta-auxiliary training stage (b), we first obtain the adapted parameters based on the auxiliary loss in the inner loop. Then, the meta-model is updated at the outer loop based on the main task loss computed on adapted parameters. At test-time (c), we simply apply the adaptation step to update the model specifically to an unseen target domain. }
\label{fig2}
\end{figure*}

\textit{Test-time adaptation/training} (TTA) aims to adapt models towards test data to overcome the performance degradation caused by distribution shifts. \cite{sun2020test} employ rotation prediction to update the model during inference. \cite{chi2021test,liu2022towards,liu2023meta} reconstruct the input images to achieve internal-learning. ARM \cite{zhang2021adaptive} incorporates test-time adaptation with DG, which meta-learns a model that is capable of adapting to unseen target domains. Meta-DMoE \cite{zhong2022meta} suggests treating each source domain as an expert, addressing domain shift by querying the relevant knowledge between the target domain data and these experts. However, those methods do not explicitly identify which knowledge and how they should be learned to enlarge the performance gain. 


\paragraph{Batch normalization.} \cite{nado2020evaluating} introduce prediction-time BN, utilizing test batch statistics for standardization. Similarly, \cite{du2020metanorm} and \cite{hu2021mixnorm} update the test statistic using predefined hyper-parameter and moving average, respectively. Conversely, \cite{schneider2020improving} propose to adapt BN statistics by combining source and test batch statistics to mitigate covariate shift.
\cite{lim2023ttn} interpolate the statistics by adjusting the importance between source and test batch statistics according to the domain-shift sensitivity of each BN layer. The primary difference with existing methods is that instead of learning a perturbable statistic, we focus on investigating the role of affine parameters towards the generalizability of few-shot learners under TT-DA setting.

\paragraph{Meta-learning.} Existing meta-learning methods can be categorized into: 1) model-based~\cite{bateni2020improved}; 2) optimization-based~\cite{ravi2017optimization}; and 3) metric-based ~\cite{snell2017prototypical}. Typical meta-learning methods utilize bi-level optimization to train a model that is applicable for downstream adaptations. Our work is built upon MAML~\cite{finn2017model}, which trains a model initialization through episodes of tasks for fast adaptation via gradient updates. Such learning paradigm has been widely applied in different vision tasks, such as zero-shot learning~\cite{wu2023metazscil} and class incremental learning ~\cite{chi2022metafscil,wu2023metagcd,wang2023triple}. In our case, the adaptation is achieved in an unsupervised manner, and the bi-level optimization is utilized to adapt to unseen domain and generalize well across all data samples in that domain.

\section{The Proposed Method}

\paragraph{Problem setting.} In this work, we consider the setting in~\cite{zhong2022meta} which we refer to as \emph{test-time domain adaptation (TT-DA)}. During the offline learning stage, we have access to $M$ labeled source domains denoted as $\{\mathcal{D}_S^i\}_{i=1}^M$. Each source domain $\mathcal{D}_S^i$ contains a set of label data, i.e. $\mathcal{D}_S^i=(x_S, y_S)^i$ where $(x, y)$ indicate the input image and corresponding label, respectively. After offline training with these source domains, we obtain a trained model and an adaptation mechanism. During testing, we are given a new target domain $D_T$. Our goal is to adapt the trained model to this target domain using only a small number of unlabeled images from the target domain. Here we assume all the domains share the same label space, but there can be a domain shift between any of these source and target domains $\{D_S^1, D_S^2, ..., D_S^M, D_T\}$. 

TT-DA is related to domain adaptation, but there are some key differences. Traditional unsupervised domain adaptation (UDA) assumes access to both source (either single or multi-source) and target domains during training. In contrast, during offline training, TT-DA does not have access to the target domain. During testing, the adaptation of TT-DA cannot access the source domains. Another related setting is source-free domain adaptation (SFDA). The key difference is that the adaptation in SFDA assumes access to a large number of unlabeled data from the target domain. In contrast, TT-DA only requires a small number of target images for the adaptation. TT-DA is a more realistic setting for many real-world scenarios. For example, in wild animal monitoring applications, after we install a surveillance camera in a new location (i.e. domain), we may want to deploy an adapted model after collecting only a few images from the new camera.
TT-DA is also related to test-time adaptation (TTA). The key difference is that TTA usually adapts the model for a batch of test examples, then the adapted model is used for prediction on the same batch of examples. In practice, it may not be realistic to adapt the model every time before a prediction is needed. In contrast, TT-DA only adapts the model once using a small batch of images. The adapted model is then used for prediction for all test images of the target domain. In real-world scenarios, this is a more realistic setting.

\subsection{Motivations}
Adapting the model using a few unlabeled data is challenging and the problem is particularly acute when unknown distribution is encountered. Within this complex setting, two fundamental questions require careful consideration: 1) What type of knowledge is most efficacious for adapting to an unseen domain? 2) How can one procure adequate supervision to guide the model's update toward that domain?

Prior work has shown that label and domain knowledge are distinctively encoded in the weight matrix and BN layers, respectively~\cite{li2016revisiting}. In the context of TT-DA, where all domains share the same label space, it implicitly indicates that label information can be learned from extensive source data~\cite{li2016revisiting}. Consequently, our approach focuses on selectively modulating the BN layers, leaving the well-acquired label knowledge undisturbed. 
Given a batch of feature map $\textbf{F}$, BN layer performs normalization and affine transformation separately on every channel as:
\begin{equation}
    \textbf{F}' = \gamma \hat{\textbf{F}} + \beta, \quad \text{where}, \quad \hat{\textbf{F}} = \frac{\textbf{F} - \mu}{\sqrt{\sigma^2 + \epsilon}}.
\end{equation}
\noindent $\mu$ and $\sigma$ are the calculated mean and variance of the batch, and $\epsilon$ is a small number to prevent division by 0. $\gamma$ and $\beta$ are the two affine parameters. During training, running mean $\mu'$ and variance $\sigma'$ are updated by each batch, and converge to the true statistics of source data. They will be adopted at the inference. However, estimating the statistics of unknown target domains using a few examples is unstable, as the data points are sparsely sampled.
To reduce the interference from such instability, we propose to leverage the $\mu'$ and $\sigma'$ computed from source data and only update $\gamma$ and $\beta$. Although the first normalization step will transform the input features towards source distribution, $\gamma$ and $\beta$ are optimized to adapt and pull the normalized feature towards target distribution. This approach is straightforward to implement and keeps the minimum number of learnable parameters without introducing any extra inference cost.

To provide supervision for $\gamma$ and $\beta$, and in the meantime, further strengthen the domain knowledge learning, we propose to utilize class-independent self-supervised learning. Specifically, an auxiliary branch is integrated in parallel with the main classification branch as shown in Fig.~\ref{fig2}(a). These two branches share the same input feature which is encoded by the backbone. The architecture of the auxiliary branch depends on the SSL algorithm -- it can be a single MLP or a more complicated structure. In this work, we do not aim to design novel SSL methods, but rather adopt the existing ones, e.g. BYOL~\cite{grill2020bootstrap}. After adapting to the target domain, the auxiliary branch can be discarded and only the original network, e.g. ResNet, is retained for inference. 

\subsection{Learning label-dependent representation} 
Given that the weight matrix encodes rich label information, and all the domains share the same label space, we first perform large-scale training on source data. We mix all the data in $\{\mathcal{D}_S^i\}_{i=1}^M$ to uniformly sample the mini-batches. The auxiliary and main branches are updated by optimizing the joint loss:
\begin{equation}
    \mathcal{L}_{Joint} = \mathcal{L}_{CE} + \lambda \mathcal{L}_{SSL}.
    \label{eq:joint}
\end{equation}
\noindent $\mathcal{L}_{SSL}$ and $\mathcal{L}_{CE}$ denote the self-supervised loss and supervised cross-entropy loss, respectively. Note that $\mathcal{L}_{CE}$ can be replaced accordingly (e.g. MSE loss) for regression problems. These two losses are balanced by $\lambda$. The running mean $\mu'$ and variance $\sigma'$ statistics collected during the training converge to the accurate statistics of source data.

\subsection{Learning to adapt to unseen domain knowledge}
The model optimized by $\mathcal{L}_{joint}$ is not necessarily prepared to adapt to unseen domains as the auxiliary loss ($\mathcal{L}_{SSL}$) and the main loss ($\mathcal{L}_{CE}$) are mutually independent. In other words, the parameters updated by the gradient from the auxiliary branch cannot guarantee positive improvement to main task. Moreover, the model lacks awareness of its subsequent learning duty, which involves adapting to unseen domains. 
\paragraph{Meta-auxiliary training.} To address such an issue, we propose a meta-auxiliary learning scheme to align the gradients between two losses and endow the model to learn to adapt to unseen domains. At the meta-auxiliary training stage, we freeze the weight matrix to preserve the rich label information. We also directly adopt the running mean $\mu'$ and variance $\sigma'$ from the source data to reduce the interference on domain information learning caused by unstable few-shot data. As a result, only the affine parameters 
$(\gamma,\beta) =\{(\gamma,\beta)^S, (\gamma,\beta)^A\}$ are learnable during this stage. Here the superscripts $S$ and $A$ are used to denote the parameters for the shared backbone and the auxiliary branch, respectively. Note that we adopt a one-layer MLP as the classification head without BN, and $(\gamma,\beta)^A$ are SSL-dependent. If the auxiliary branch has no BN, it can be simply ignored.

To train $(\gamma,\beta)$, we employ episodic learning as in meta-learning~\cite{finn2017model} to consider every domain as a ``task''. We perform nested optimization at the task level instead of instance level, so that $(\gamma,\beta)$ are meta-trained to fulfill the task of learning to adapt to unseen domains. For each iteration, we select a source domain (task) $\mathcal{D}_S^i$. We sample an unlabeled support set $\mathcal{S}^i$ and a labeled query set $\mathcal{Q}^i$ from $\mathcal{D}_S^i$. We first adapt $(\gamma,\beta)$ to this domain using SSL on the support set $\mathcal{S}^i$ with a learning rate of $\alpha$ as:
\begin{equation}
(\tilde{\gamma},\tilde{\beta}) = (\gamma,\beta) -\alpha\nabla_{(\gamma,\beta)}\mathcal{L}_{SSL}(\mathcal{S}^i;\theta,(\gamma,\beta)),
\label{eq:inner}
\end{equation}
where $\theta$ denotes all the weight matrices. Ideally, $(\tilde{\gamma},\tilde{\beta})$ should adapt to the domain $\mathcal{D}_S^i$ and improve the main branch. In other words, the adapted model should generalize to all data in that domain in terms of the main task. Therefore, we evaluate $(\tilde{\gamma},\tilde{\beta})$ on the disjoint labeled query set $\mathcal{Q}^i$ using Eq.~\ref{eq:joint} as the meta-objective and update as:
\begin{equation}
(\gamma,\beta) \leftarrow (\gamma,\beta) -\delta\nabla_{(\gamma,\beta)} \mathcal{L}_{Joint}(\mathcal{Q}^i;\theta, (\tilde{\gamma},\tilde{\beta})),
\label{eq:outer}
\end{equation}
where $\delta$ is the outer loop learning rate. Updating $(\gamma,\beta)^A$ requires supervision from $\mathcal{L}_{SSL}$, therefore, we use the joint loss.
Note, in Eq.~\ref{eq:outer}, the evaluation is conducted on $(\tilde{\gamma},\tilde{\beta})$ but the gradient update is carried out on original $(\gamma,\beta)$ to achieve meta-level updates. The symbol for input and output $(x, y)$ is omitted in Eq.~\ref{eq:inner} and Eq.~\ref{eq:outer} for simplicity and the process is repeated until converged. Alg.~\ref{Algorithm one} and Fig.~\ref{fig2}(b) elaborate the whole training pipeline. 

\paragraph{Meta-auxiliary testing.} The meta-parameters $(\gamma,\beta)$ have been learned specifically to facilitate the domain knowledge adaptation on unseen target domains. At test-time, given an unseen target domain $\mathcal{D}_T$, the adapted parameters $(\tilde{\gamma},\tilde{\beta})$ are obtained by simply performing the inner loop as Line 12 in Alg.~\ref{Algorithm one} and Fig.~\ref{fig2}(c) using the support set consisting of a few unlabeled images. Then the adapted model can perform inference on all test samples in that domain, and the auxiliary branch is discarded to retain the computational cost.



\begin{algorithm}[t]
\caption{Meta-auxiliary training of MABN}
\begin{algorithmic}
\label{Algorithm one}
\REQUIRE $\alpha$, $\delta$, $\eta$: learning rates; $B$: meta batch size \\
\REQUIRE $\{\mathcal{D}^i_{S}\}^M_{i=1}$: data of source domains\\
\STATE 1: Initialize weight matrix $\theta$ and affine params $(\gamma,\beta)$ \\
\STATE 2: // Learning label representation on mixed source data \\
\STATE 3:  $(\theta, \gamma,\beta) \leftarrow (\theta, \gamma,\beta) -\eta\nabla_{(\theta, \gamma,\beta)} \mathcal{L}_{Joint}(\mathcal{D}_{S};\theta;(\gamma,\beta))$\\
\STATE 4: \textbf{while} not converged \textbf{do}\\
\STATE 5: \quad // Learning to adapt to domain-specific knowledge
\STATE 7: \quad Sample a meta batch of $B$ source domains: $\{\mathcal{D}^i_S\}^B_{i=1}$
\STATE 8: \quad Reset the loss of current meta batch: $\mathcal{L}_B=0$
\STATE 9: \quad \textbf{for} each $\mathcal{D}^i_S$ in $\{\mathcal{D}^i_S\}^B_{i=1}$ \textbf{do}
\STATE 10: \quad \quad Sample support and query set: $(\mathcal{S}^i$, $\mathcal{Q}^i) \sim\mathcal{D}^i_S$
\STATE 11: \quad \quad // Perform adaptation via self-supervised loss 
\STATE 12: \quad \quad $(\tilde{\gamma},\tilde{\beta})=(\gamma,\beta) -\alpha \nabla_{(\gamma,\beta)}\mathcal{L}_{SSL}(\mathcal{S};\theta;(\gamma,\beta)) $ \\
\STATE 14: \quad \quad // Evaluate the adapted $(\tilde{\gamma},\tilde{\beta})$ using $\mathcal{Q}$ and \\ 
\STATE 15: \quad \quad // accumulate the loss\\ 
\STATE 16: \quad \quad $\mathcal{L}_B=\mathcal{L}_B+\mathcal{L}_{Joint}(\mathcal{Q};\theta;(\tilde{\gamma},\tilde{\beta})^{S,A})$ \\
\STATE 17: \quad \textbf{end for}
\STATE 18: \quad  // Update $(\gamma,\beta)$ for current meta batch
\STATE 19: \quad $(\gamma,\beta) \leftarrow (\gamma,\beta) -\delta\nabla_{(\gamma,\beta)} \mathcal{L}_{B}$
\STATE 21: \textbf{end while}
\end{algorithmic}
\end{algorithm}

\section{Experiments}

\begin{table*}[htbp]
\begin{center}
\setlength{\tabcolsep}{1.23mm}{
\begin{tabular}{c|cc|c|c|cc|cc}
\toprule[1pt]
\multirow{2}{*}{\textbf{Methods}} &\multicolumn{2}{c|}{\textbf{iWildCam}} &\multicolumn{1}{c|}{\textbf{Camelyon17}} &\multicolumn{1}{c|}{\textbf{RxRx1}}  &\multicolumn{2}{c|}{\textbf{FMoW}} &\multicolumn{2}{c}{\textbf{PovertyMap}} \\ 
                         &\textbf{Acc} &\textbf{Macro F1} &\textbf{Acc} &\textbf{Acc} &\textbf{WC Acc} &\textbf{Avg Acc} &\textbf{WC Pearson r} &\textbf{Pearson r} \\ \hline
ERM &71.6$\pm$2.5 &31.0$\pm$1.3 &70.3$\pm$6.4 &29.9$\pm$0.4 &32.3$\pm$1.25 &53.0$\pm$0.55 &0.45$\pm$0.06 &0.78$\pm$0.04\\
CORAL &73.3$\pm$4.3 &32.8$\pm$0.1 &59.5$\pm$7.7 &28.4$\pm$0.3 &31.7$\pm$1.24 &50.5$\pm$0.36 &0.44$\pm$0.06 &0.78$\pm$0.05\\
Group DRO &72.7$\pm$2.1 &23.9$\pm$2.0 &68.4$\pm$7.3 &23.0$\pm$0.3 &30.8$\pm$0.81 &52.1$\pm$0.5 &0.39$\pm$0.06 &0.75$\pm$0.07 \\
IRM &59.8$\pm$3.7 &15.1$\pm$4.9 &64.2$\pm$8.1 &8.2$\pm$1.1 &30.0$\pm$1.37 &50.8$\pm$0.13 &0.43$\pm$0.07 &0.77$\pm$0.05 \\ 
ARM-CML &70.5$\pm$0.6 &28.6$\pm$0.1 &84.2$\pm$1.4 &17.3$\pm$1.8 &27.2$\pm$0.38 &45.7$\pm$0.28 &0.37$\pm$0.08 &0.75$\pm$0.04 \\
ARM-BN &70.3$\pm$2.4 &23.7$\pm$2.7 &87.2$\pm$0.9 &31.2$\pm$0.1&24.6$\pm$0.04 &42.0$\pm$0.21 &0.49$\pm$0.21 &\textbf{0.84$\pm$0.05} \\
ARM-LL &71.4$\pm$0.6 &27.4$\pm$0.8 &84.2$\pm$2.6 &24.3$\pm$0.3 &22.1$\pm$0.46 &42.7$\pm$0.71 &0.41$\pm$0.04 &0.76$\pm$0.04 \\
Meta-DMoE &77.2$\pm$0.3 &34.0$\pm$0.6 &91.4$\pm$1.5 &29.8$\pm$0.4 &35.4$\pm$0.58 &52.5$\pm$0.18 &0.51$\pm$0.04 &0.80$\pm$0.03 \\
PAIR &74.9$\pm$1.1 &27.9$\pm$0.9 &74.0$\pm$7.2 &28.8$\pm$0.0 &35.4$\pm$1.30 &- &0.47$\pm$0.09 &- \\ \hline
MABN (ours) &\textbf{78.4$\pm$0.6} &\textbf{38.3$\pm$1.2} &\textbf{92.4$\pm$1.9} &\textbf{32.7$\pm$0.2} &\textbf{36.6$\pm$0.41} &\textbf{53.2$\pm$0.52} &\textbf{0.56$\pm$0.05} &\textbf{0.84$\pm$0.04}\\
\bottomrule[1pt]
\end{tabular}}
\caption{\textbf{Comparison with the state-of-the-arts on the WILDS benchmark under the out-of-distribution setting.} Metric means and standard deviations are reported across replicates. For each comparison method, we either obtain its result directly reported in the leaderboard or use the released official model with default parameters. }
\label{tab1}
\end{center}
\end{table*}


\paragraph{Dataset and evaluation metrics.} In this work, follow Meta-DMoE~\cite{zhong2022meta} to evaluate our method on five benchmarks from WILDS~\cite{koh2021wilds}: iWildCam \cite{beery2021iwildcam}, Camelyon17 \cite{bandi2018detection}, RxRx1 \cite{taylor2019rxrx1}, FMoW \cite{christie2018functional} and PovertyMap \cite{yeh2020using}. Note, that we follow the official training/validation/testing splits, and report the same metrics as in \cite{koh2021wilds}, including accuracy, Macro F1, worst-case (WC) accuracy, Pearson correlation (r), and its worst-case counterpart. We also evaluate on DomainNet benchmark~\cite{peng2019moment}. Detailed descriptions of the benchmarks are provided in the supplement.




\paragraph{Model architectures.} For faithful comparison, we follow WILDS \cite{koh2021wilds} to use ResNet50 \cite{he2016deep}, DenseNet121\cite{huang2017densely} and ResNet18 \cite{yeh2020using} as the backbones of our proposed method for iWildCam/RxRx1, Camelyon17/FMoW and PovertyMap datasets. The output of the last average pooling layer of the backbone is served as the input of auxiliary and main branches. As for the auxiliary branch, we select BYOL~\cite{grill2020bootstrap} as the self-supervised learning method. Its architecture consists of two MLP layers with BN and ReLU activation, we refer the readers to the original paper for more details. For the main branch, we simply use a linear layer for both classification and regression tasks.

\paragraph{Implementation.} 
We follow~\cite{zhong2022meta} to use ImageNet-1K \cite{deng2009imagenet} pre-trained weights as the initialization to perform joint training. Adam optimizer is used to minimize Eq.~\ref{eq:joint} with a learning rate (LR) of $1e^{-4}$ for 20 epochs. LR is reduced by a factor of 2 when the loss reaches a plateau. $\lambda$ in Eq.~\ref{eq:joint} is set to 0.1. During meta-auxiliary training, we fix the weight matrix of the entire network, and directly use the running statistics $\mu'$ and $\sigma'$ for the BN layers. Only the affine parameters $\gamma$ and $\beta$ of the BN layers are further optimized using Alg. 1 for 10 epochs with fixed LR of $3e^{-4}$ for $\alpha$ and $3e^{-5}$ for $\delta$. During testing, for each target domain, we randomly sample 12 images for iWildCam and 32 images for the rest datasets to perform adaptation first (Line 12-13 of Alg.~\ref{Algorithm one}). The adapted model is then used to test all the images in that domain. The same process is repeated for all target domains. All the experiments are conducted with 5 random seeds to show the variation.

\subsection{Experimental Results} 
\paragraph{Main results.} We compare the proposed approach with a variety of methods showing on the WILDS leaderboard, including non-adaptive methods ERM \cite{vapnik1999overview}, CORAL \cite{sun2016deep}, Group DRO \cite{sagawa2019distributionally}, IRM \cite{arjovsky2019invariant} and adaptive methods ARM \cite{zhang2021adaptive}, Meta-DMoE \cite{zhong2022meta}, PAIR \cite{Chen2023Pareto}. The experimental results in Tab.~\ref{tab1} show that the proposed method outperforms all other methods under 8 evaluation metrics across all five datasets. Specifically, our MABN surpasses the most recent method PAIR in classification accuracy by 3.5\%, 18.4\% and 3.9\% on iWildCam, Camelyon17 and RxRx1 datasets. We also outperform Meta-DMoE by a (\%) margin of 1.2/4.6 on iWildCam, 1.0 on Camelyon17, 2.9 on RxRx1, 1.2/0.7 on FMoW and 0.05/0.04 on PovertyMap. Note, our method only uses a single model which is more lightweight while Meta-DMoE has a group of teacher models. These results show the effectiveness of identifying the crucial parameters for adapting to domain-specific knowledge. Such adaptability is important for enhancing the generalization of the target domain. Tab.~\ref{tab:domainet} reports the superiority of our method on DomainNet.

\begin{table}[!th]
\centering
\setlength{\tabcolsep}{1.2mm}
\small
\begin{tabular}{cccccccc}
\toprule
\textbf{Method} &\textbf{clip} &\textbf{info} &\textbf{paint} &\textbf{quick} &\textbf{real} &\textbf{sketch} &\textbf{avg} \\ \midrule
ARM &49.7 &16.3 &40.9 &9.4 &53.4 &43.5 &35.5 \\
Meta-DMoE &63.5 &21.4 &51.3 &14.3&62.3 &52.4 &44.2 \\
Ours      &\textbf{64.2} &\textbf{23.6} &\textbf{51.5} &\textbf{15.2} &\textbf{64.6} &\textbf{54.1} &\textbf{45.5}     \\ \bottomrule
\end{tabular}
\caption{\textbf{Comparison on the DomainNet (Accuracy).}}
\label{tab:domainet}
\end{table}

\begin{table}[t]
\begin{center}
\setlength{\tabcolsep}{1.2mm}{
\begin{tabular}{c|ccc}
\toprule[1pt]
\textbf{ Adapted} $(\tilde{\gamma}, \tilde{\beta})$ &\textbf{No adapt} &\textbf{Not-matched} &\textbf{Matched} \\ \hline
Accuracy &74.69 &72.39 &\textbf{78.40}  \\
Macro-F1 &36.77 &33.32 &\textbf{38.27} \\
\bottomrule[1pt]
\end{tabular}}  
\caption{\textbf{Verification of domain knowledge learning.} ``No adapt" means the meta-learned $(\gamma, \beta)$ are used for all domains without adaptation. ``Not matched" means each target domain randomly uses the adapted $(\tilde{\gamma}, \tilde{\beta})$ from other domains instead of its own. ``Matched" means each target domain uses its own adapted $(\tilde{\gamma}, \tilde{\beta})$.
}
\label{domain_info}
\end{center}
\end{table}

\begin{figure}[t]
\centering
		\label{fig:example}
		\begin{subfigure}{.20\textwidth}
			\includegraphics[width=\textwidth]{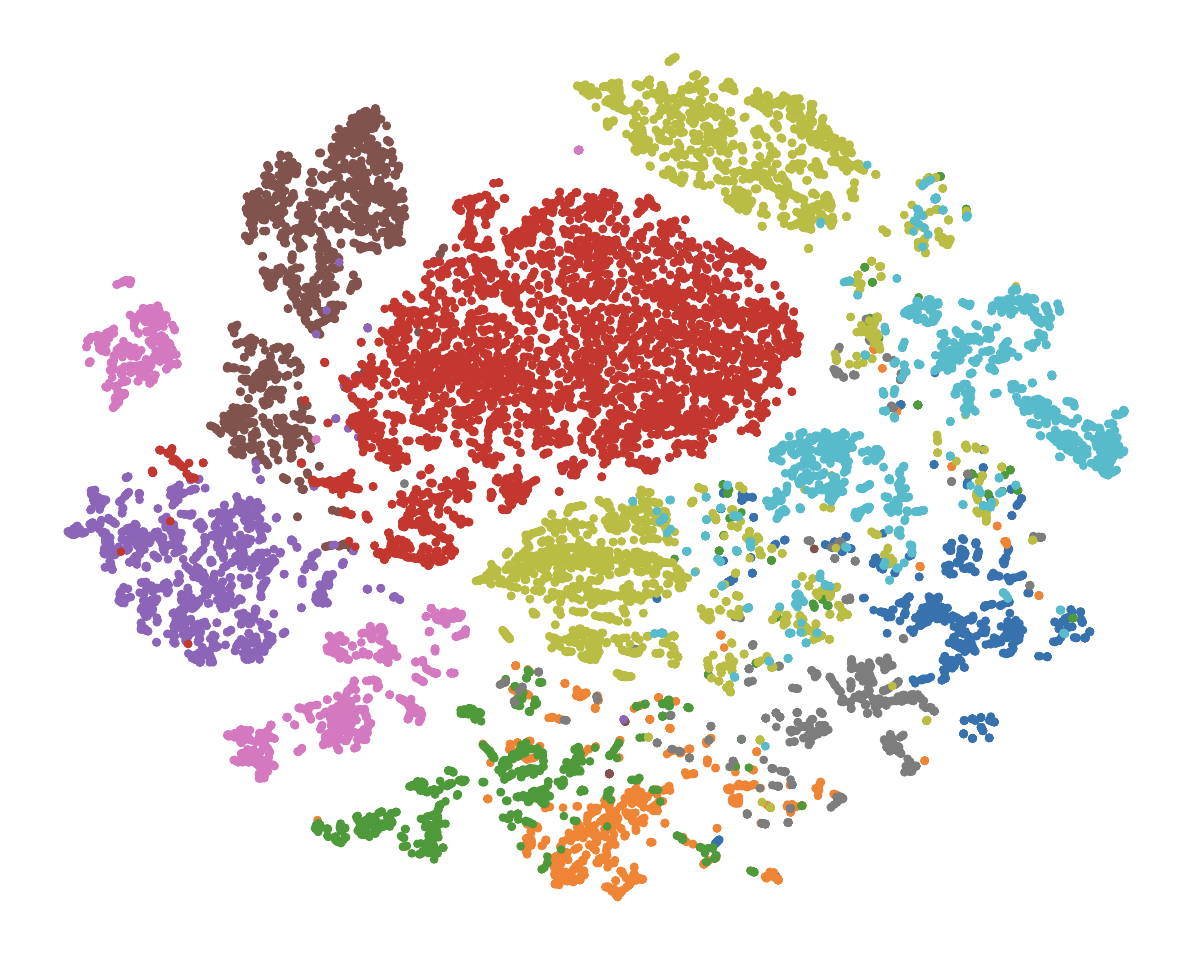}
			\caption{Before adaptation}
		\end{subfigure}
		\begin{subfigure}{.20\textwidth}
			\includegraphics[width=\textwidth]{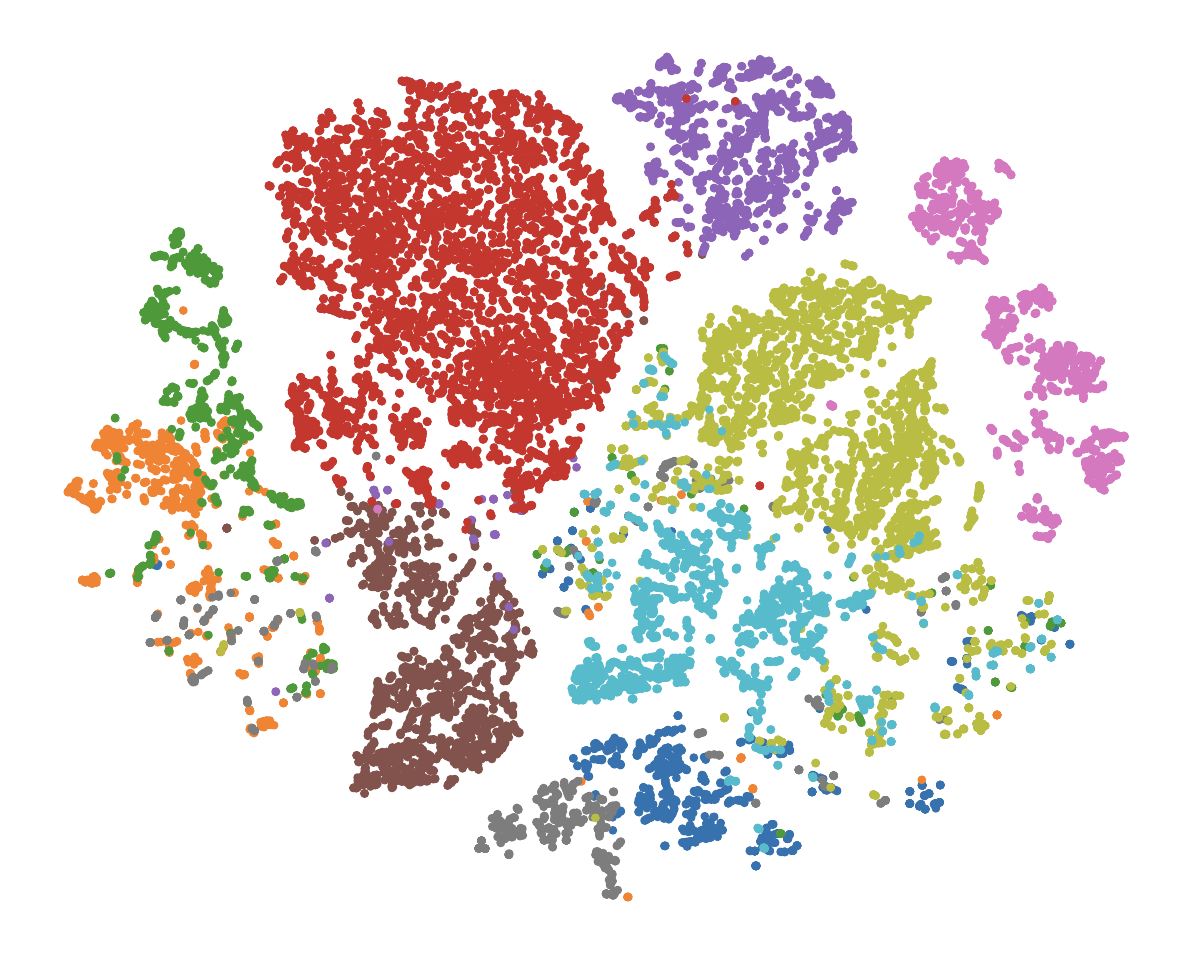}
			\caption{After adaptation}
		\end{subfigure}
\caption{\textbf{t-SNE visualization of features before and after adaptation.} Each data sample is represented as a point, and each color corresponds to a class randomly selected from the target domain of the iWildCam dataset.}
\label{tsne}
\end{figure}

\paragraph{Is MABN really learning the domain knowledge?} 
To show that our adapted models have learned domain-specific information and adapt to each of the target domains, we conduct an experiment to shuffle the adapted $(\tilde{\gamma}, \tilde{\beta})$ across target domains. Specifically, assume there are $N$ target domains, we save all the adapted affine parameters as $\{(\tilde{\gamma}, \tilde{\beta})_j\}_{j=1}^N$. For $j^{th}$ target domain, we randomly use $(\tilde{\gamma}, \tilde{\beta})_u$ which belongs to other domains (i.e., $i \neq u$), we name it ``Not matched". 
We then compare it with when the $j^{th}$ target domain utilizes its own $(\tilde{\gamma}, \tilde{\beta})_j$, which we refer to as ``Matched".
As reported in Tab.~\ref{domain_info}, several observations can be made: 1) when the target domains utilize non-matched adapted domain knowledge, the performance dropped significantly. However, the performance boosts when the target domain uses its own knowledge; 2) Using the correct adapted $(\tilde{\gamma}, \tilde{\beta})$, immense improvement is made compared to the non-adaptive model. Thus, we can conclude that our method learns unique domain knowledge for every target domain and improves the overall performance. We also visualize the features before and after adaptation using t-SNE \cite{van2008visualizing}. As shown in Fig.~\ref{tsne}, each class cluster is more discriminative after adaptation.

\begin{table}[t]
\begin{center}
\setlength{\tabcolsep}{0.9mm}{
\begin{tabular}{c|cc|cc}
\toprule[1pt]
\multirow{2}{*}{\textbf{Method}} &\multicolumn{2}{c|}{\textbf{Update BN}} &\multicolumn{2}{c}{\textbf{Update Affine}} \\ \cline{2-5}
               &\textbf{Acc}  &\textbf{Macro F1} &\textbf{Acc} & \textbf{Macro F1}  \\\hline
TENT (min. entropy) &33.27 &0.77 &75.92 &36.40  \\
Our (min. auxiliary) &75.86 &36.76 &78.40 &38.27 \\
Our+TENT &75.84 &31.93 &79.68 &38.85 \\
\bottomrule[1pt]
\end{tabular}}
\caption{\textbf{Comparison and integration of our MABN with other TTA method.} TENT learns more label-related knowledge by minimizing entropy, which can be seamlessly combined with our method to achieve superior performance. }
\label{tent}
\end{center}
\end{table}

To further show that our method is learning domain knowledge, we seamlessly integrate MABN with an entropy-based TTA method TENT~\cite{wang2020tent}, which is more label-related. To do so, for each target domain, we first use MABN to adapt the domain and then apply TENT on each batch. We also compare the case of taking the whole BN layer and only the affine parameters. Several conclusions can be drawn from Tab.~\ref{tent}: 1) Adapting the entire BN layer proves to be problematic, as the normalization is unstable and interferes with the affine parameters; 2) Our method outperforms TENT due to the fact that label-related information is already contained within the weight matrix, thereby necessitating the learning of domain-specific information;
3) Since we aim to learn the domain information, integrating with TENT further pushes the boundary by a large margin (1.28\% and 0.58\% on accuracy and F1 score).

\begin{table}   
\begin{center}
\setlength{\tabcolsep}{1.0mm}{
\begin{tabular}{c|cccc|cc}
\hline
\multirow{2}{*}{\textbf{Index}} &\multirow{2}{*}{$~\bm{\bm{SSL}}~$} &\multirow{2}{*}{$\bm{\bm{Param.}}$} &\multirow{2}{*}{ $\bm{\bm{TS}}$} &\multirow{2}{*}{$\bm{\bm{Adapt}}$} &\multicolumn{2}{c}{\textbf{iWildCam}} \\
 & & & &  &\textbf{Acc} & \textbf{F1}\\ \hline \hline
\rule{0pt}{8pt} 1 &\ding{55} &All &CE &\ding{55}  &68.7 &31.3 \\ 
\rule{0pt}{8pt} 2 &\ding{51} &All &Joint &\ding{55}  &70.5 &33.2 \\ 
\rule{0pt}{8pt} 3 &\ding{51} &BN &Joint &\ding{51}  &68.2 &30.5 \\ 
\rule{0pt}{8pt} 4 &\ding{51} &Aff &Joint &\ding{51}  &71.1 &33.9 \\ 
\rule{0pt}{8pt} 5 &\ding{51} &All &Meta &\ding{51}  &72.0 &29.4 \\ 
\rule{0pt}{8pt} 6 &\ding{51} &Aff &Meta &\ding{55} &74.7 &36.8 \\
\rule{0pt}{8pt} 7 &\ding{51} &Aff &Meta &\ding{51} &78.4 &38.3 \\ \hline
\end{tabular}}
\caption{\textbf{Ablation studies for different components of our framework.} ${SSL}$ denotes SSL branch. ${Param.}$ denotes which parameters are updating, including the whole network (``All"), BN layer (``BN") or only the affine parameters (``Aff"). ${TS}$ denotes the training scheme. ${Adapt}$ denotes whether adapting to each target domain.}  
\label{ablation_studies}
\end{center}
\end{table} 

\paragraph{Ablation studies.} To evaluate the contribution from each component in MABN on the iWildCam dataset. As reported in Tab.~\ref{ablation_studies}, several conclusions can be drawn: 1) (Index 1 vs. 2) adding the SSL branch and jointly training is beneficial to handle distribution shift. The hypothesis is that SSL improves the discrimination of the learned feature; 2)  (Index 2 vs. 3) adapting the model by updating the whole BN even degrades the performance as the normalization statistic learned from a few images is unstable. However, freezing the normalization and only updating the affine parameters overturns the results (Index 4) and positive improvement is observed. The results confirm our hypothesis: the unstable normalization statistics can interfere the adaptation process and the affine parameters alone are more effective to adapt to unseen domains; 3) Our meta-auxiliary training aims to meta-learn the affine parameters that can adapt to any domains effectively. It implicitly indicates that the meta-learned affine parameters are more general to all distributions instead of the one only fit to source domains. Therefore, even without adaptation, the meta-learned model achieves better generalization (Index 4 vs. 6). Besides, our meta-objective aligns two learning objectives, so that the adapted affine parameters by SSL to target domains further push the performance boundary (Index 7); 4) But without properly disentangling the learning knowledge and identifying the adaptive parameters, 
it is ineffective to adapt to unseen domains (Index 5).

\paragraph{Hyper-parameter analysis.} We study the performance of our MABN under various parameter settings. We first investigate the sensitivity on the number of unlabeled data used for adaptation at test-time. Intuitively, more unlabeled data from a target domain reveals a more precise representation of its underlying distribution. Consequently, as reported in Fig. 6 in the supplementary, increasing the number of unlabeled data empowers MABN to more effectively exploit the domain-specific knowledge. Moreover, in the extreme case with only one unlabeled data, MABN still performs better than without adaptation. These results again indicate that adapting with only affine parameters is more effective by creating less ambiguity from unstable statistics. We also evaluate MABN with different SSL tasks as in Tab.~\ref{ssl}. It is observed that different SSL tasks affect MABN. However, they are all superior over the baseline. On the other hand, our meta-auxiliary learning scheme is able to improve over joint training for all SSL methods even with ViT~\cite{dosovitskiy2020image}, in which the last layer is updated as it does not have BN layers. It is highlighted that the proposed MABN is general and serves as a plug-and-play module for existing methods to improve generalization.

\begin{table}[t]
\begin{center}
\setlength{\tabcolsep}{0.05mm}{
\begin{tabular}{c|c|c|cc}
\toprule[1pt]
\multirow{2}{*}{\textbf{Self-supervised}} &\multirow{2}{*}{\textbf{Backbone}} &\multirow{2}{*}{\textbf{Training}} &\multicolumn{2}{c}{\textbf{iWildCam}} \\ \cline{4-5}
               &  & & \textbf{Acc} & \textbf{F1}  \\\hline
None (baseline) &ResNet50 &CE &68.7 &31.3  \\
Rotation \cite{sun2020test} &ResNet50 &Joint &69.2 &31.5  \\
Rotation \cite{sun2020test} &ResNet50 &Meta &72.8 &33.0 \\
MAE \cite{he2022masked} &ViT-Base &Joint &71.7 &33.8  \\
MAE \cite{he2022masked} &ViT-Base &Meta &74.9 &35.1  \\\hline\hline
Ours (BYOL)  &ResNet50 &Joint &70.5 &33.2  \\
Ours (BYOL) &ResNet50 &Meta &78.4 &38.3 \\
\bottomrule[1pt]
\end{tabular}}
\caption{\textbf{Evaluation on different SSL tasks.} Note, we adapt ViT-Base by updating the last block as it does not have BN layer, verifying the effectiveness of meta-auxiliary training.}
\label{ssl}
\end{center}
\end{table}

\section{Conclusion}
In this paper, we have presented a simple yet effective method MABN for test-time domain adaptation. Specifically, our proposed method achieves domain knowledge learning by only adapting the affine parameters in BN. To further extract domain knowledge from a limited amount of unlabeled data, we establish an auxiliary branch with label-independent self-supervised learning. In addition, we introduce a meta-learning-based optimization strategy so that the affine parameters are learned in a way that facilitates domain knowledge adaptation. Extensive experiments on various real-world domain shift benchmarks demonstrate the superiority of MABN over the state-of-the-art.

\section{Acknowledgments}
This work was supported by the National Key Research and Development Project (No. 2018AAA0100300),  the Fundamental Research Funds for the Central Universities (No.2022JBZY019). Yang Wang acknowledge the funding support of an NSERC Discovery grant and the Gina Cody Research and Innovation fellowship.

\bibliography{aaai24}

\newpage

\setcounter{figure}{3}
\setcounter{table}{5}
\section{Appendix}
\subsection{Elaboration on Batch Normalization (BN)}
In the main paper (motivation section), we discussed our motivation on regarding the BN. Eq. 1 in the main paper is simplified. In this supplementary material, we elaborate more about the BN operation. For each layer, given a feature map $\textbf{F}$ with dimension $\mathbb{R}^{b\times s \times c}$, where $b$ is the batch size, $s$ is the spatial dimension, $c$ is the channel number. BN first normalize each channel $j\in\{1,...,c\}$ as:
\begin{equation}
    \hat{\textbf{F}}_j = \frac{\textbf{F}_j - \mu_j}{\sqrt{\sigma^2_j + \epsilon}},
\end{equation}
where $\mu_j$ and $\sigma_j$ are the mean and variance of $j^{th}$ channel. The normalized feature is then further transformed by two affine parameters as:
\begin{equation}
    \textbf{F}'_j = \gamma_j \hat{\textbf{F}}_j + \beta_j,
\end{equation}
where $\gamma_j$ and $\beta_j$ are the scaling and shifting parameters. Such BN operation ensures that the distribution of the features is unchanged across batches when they are sent to the next weight matrix. During training, the moving average statistics $\mu'_j$ and $\sigma'_j$ are updated as:
\begin{equation}
    \mu'_j = m\times \mu'_j + (1-m)\times \mu_j,
\end{equation}
\begin{equation}
    \sigma'_j = m\times \sigma'_j + (1-m)\times \sigma_j,
\end{equation}
where $m$ is the momentum, meaning the moving average is decayed by $m$ but updated by a small amount on $\mu_j$ or $\sigma_j$. 
With many iterations, $\mu_j$ or $\sigma_j$ gradually converge to stable numbers that fit the distribution of the source domains. And $\mu_j$ or $\sigma_j$ will be used for every data at inference.
Obviously, such normalization statistics and learned affine transformation are only tailored for data that is close to the distribution of source data. Once, domain shift occurs, BN could be problematic. 

In the setting of TT-DA, for each target domain, we are allowed to update the model using a few unlabeled data. The number is much smaller than the training batch size. During training, it is common practice to set a large batch size (e.g. 128 or 256) to stabilize the training. However, the images for adaptation are way less than that (e.g. 12 images). It causes the problem of estimation of the normalization statistics, causing the interference of learning the target distribution. Therefore, it motivates us to fix the normalization step, and only update the learnable affine parameters.

\begin{figure}[H]
\centering
\centerline{\includegraphics[scale=0.59]{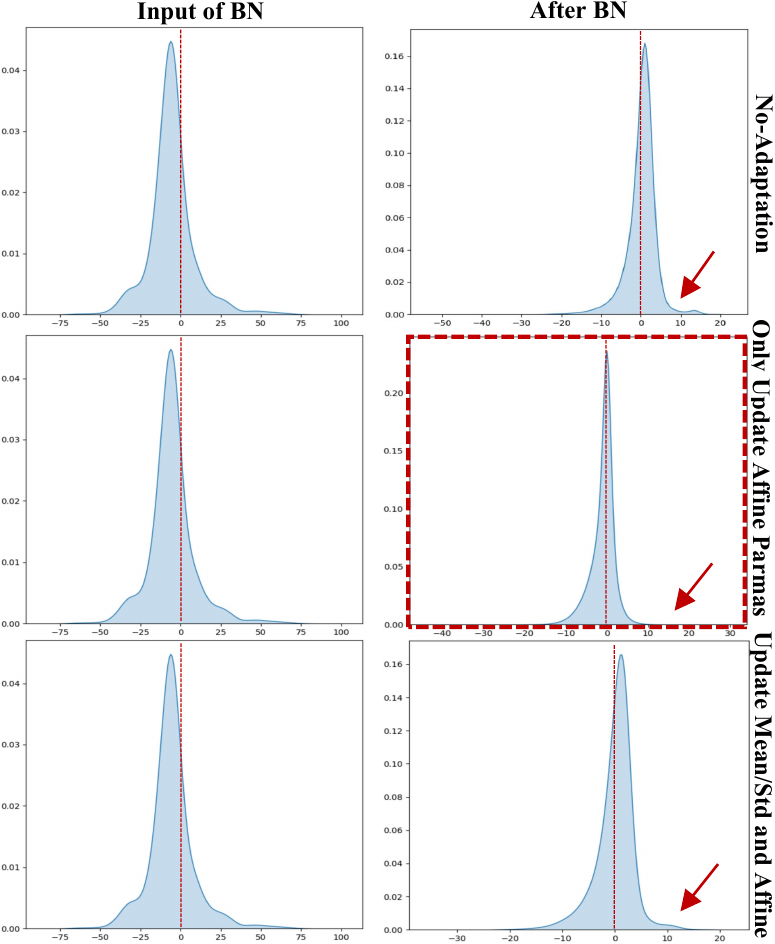}}
\caption{\textbf{Illustration of the intermediate features before and after BN for each adaptation method.} We randomly select a layer and a channel for illustration. The first column shows the distribution of the feature before BN layer and the second column shows the feature after BN.} 
\label{fig1_1}
\end{figure}

\begin{figure}[h]
\centering
\centerline{\includegraphics[scale=0.18]{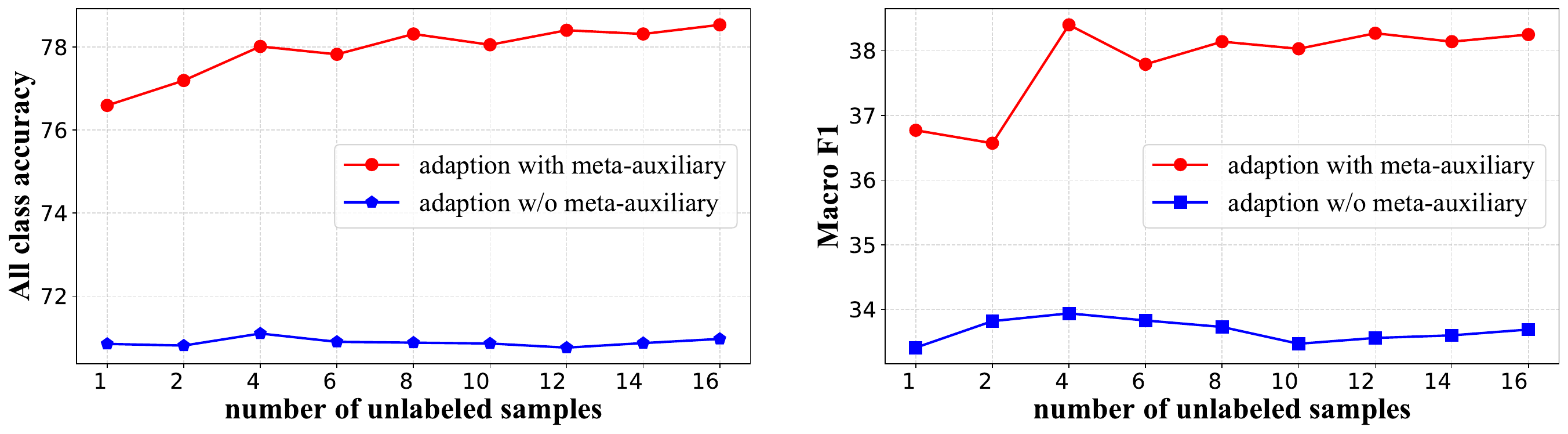}}
\caption{\textbf{Adaptive evaluation on various number of unlabeled samples without meta-auxiliary training}.} 
\label{fig2_1}
\end{figure}

\begin{table*}[!h]
\centering
\begin{tabular}{cccccccc}
\toprule
\textbf{Method} &\textbf{clip} &\textbf{info} &\textbf{paint} &\textbf{quick} &\textbf{real} &\textbf{sketch} &\textbf{avg} \\ \midrule
ARM &49.7(0.3) &16.3(0.5) &40.9(1.1) &9.4(0.1) &53.4(0.4) &43.5(0.4) &35.5 \\
Meta-DMoE &63.5(0.2) &21.4(0.3) &51.3(0.4) &14.3(0.3) &62.3(1.0) &52.4(0.2) &44.2 \\
Ours      &\textbf{64.2(0.3)} &\textbf{23.6(0.4)} &\textbf{51.5(0.2)} &\textbf{15.2(0.3)} &\textbf{64.6(0.5)} &\textbf{54.1(0.4)} &\textbf{45.5}     \\ \bottomrule
\end{tabular}

\caption{\textbf{Comparison on the DomainNet with std across 3 random seeds.}}
\label{tab:domainet_1}
\end{table*}
\begin{figure}[h]
\centering
\centerline{\includegraphics[scale=0.18]{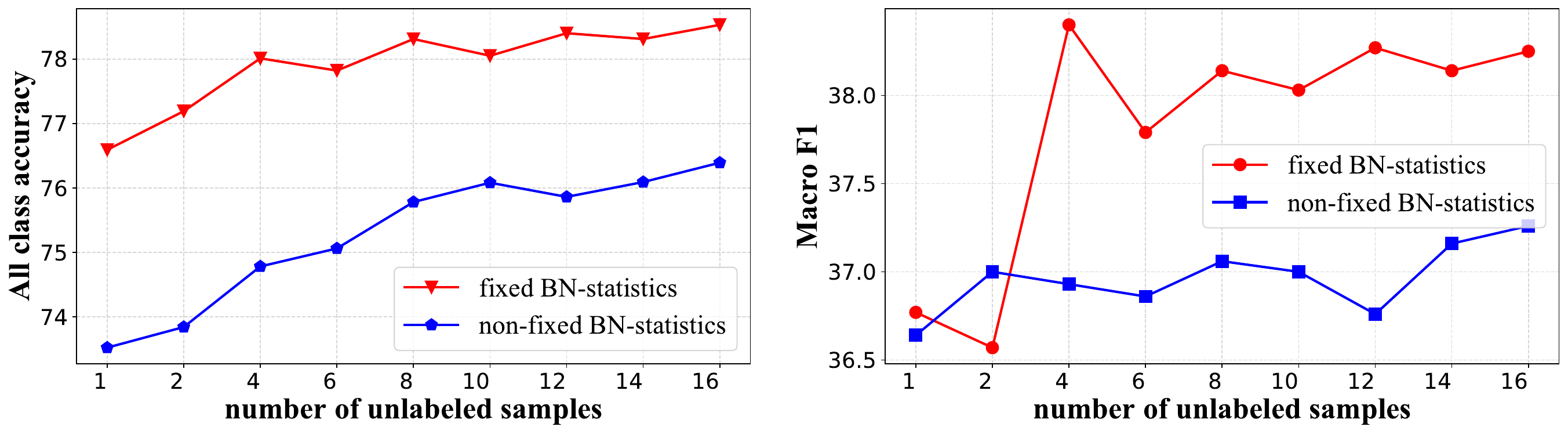}}
\caption{\textbf{Evaluation on various number of unlabeled samples for adaptation}. We report both fixed and non-fixed BN statistics cases.} 
\label{unlabeled samples}
\end{figure}

We visualize the distribution of feature before and after BN for different adaptation methods as in Fig.~\ref{fig1_1}. The right of first row shows that without adaptation, there is distortion at the tail of the distribution. The right of third row shows that when updating both normalization statistics and affine parameters, the distortion still exists and the distribution is no longer zero-centered. The right of second row shows that with only adapting the affine parameters, the distortion at the tail disappeared and the whole distribution is more centered at 0. It indicates that, when keeping the normalization statistics fix, learning the distribution is more effective with less interference.

\subsection{Adaption without Meta-Auxiliary Learning}
The main contribution of meta-auxiliary learning is to align the two learning objectives so that the model updated by the auxiliary branch can benefit the main classification/regression branch. To further show its effectiveness, we directly use the jointly trained model for adaptation. Fig.~\ref{fig2_1} shows that the performance is quite stable regardless of how many images are used for adaptation. Sometimes, the performance even drops compared to the baseline without adaptation. It indicates the importance of enforcing the two objectives to be consistent and mutually dependent.

\subsection{Details of Datasets}
\noindent\textbf{WILDS: } We evaluate on 5 image testbeds from WILDS: iWildCam consists of 203,029 images from 323 camera traps, with 243 and 48 camera traps serving as source and target domains. Camelyon17 comprises 455,954 tissue patches from 5 hospitals, where 3 of them are source domains, and 1 hospital is the target domain. RxRx1 includes 125,510 cell images from 51 running batches, with 33 and 14 batches considered as source and target domains. In PovertyMap, there are 19,669 satellite images taken in 46 countries, with 26 and 10 of them considered as source and target domains. FMoW contains 141,696 satellite images taken in 5 geographical regions and at 16 different times. FMoW has 55 source domains and 10 target domains. The best model is selected with the highest validation performance.

\noindent\textbf{DomainNet:} contains 6 domains with 569K images of 345 classes. We follow the official leave-one-domain-out to train 6 models. Note, that we follow the official training/testing splits instead of randomly selecting a portion of the dataset as a test split.

\end{document}